\documentclass[10pt,twocolumn,letterpaper]{article}

\usepackage{wacv}      

\usepackage{graphicx}
\usepackage{amsmath}
\usepackage{amssymb}
\usepackage{booktabs}
\usepackage{float}
\usepackage{multirow}
\usepackage{booktabs}

\usepackage[pagebackref,breaklinks,colorlinks]{hyperref}

\usepackage[capitalize]{cleveref}
\crefname{section}{Sec.}{Secs.}
\Crefname{section}{Section}{Sections}
\Crefname{table}{Table}{Tables}
\crefname{table}{Tab.}{Tabs.}

\def\wacvPaperID{654} 
\def\confName{WACV}
\def\confYear{2024}

\begin{document}

\title{PlantPlotGAN: A Physics-Informed Generative Adversarial Network for \\ Plant Disease Prediction}

\author{Felipe A. Lopes\textsuperscript{1,3}, Vasit Sagan\textsuperscript{1,*}, Flavio Esposito\textsuperscript{2}\\
\textsuperscript{1}Taylor Geospatial Institute, St. Louis, MO 63108, USA\\
\textsuperscript{2}Department of Computer Science, Saint Louis University, St. Louis, MO 63108, USA\\
\textsuperscript{3}Federal Institute of Alagoas, Arapiraca, Brazil\\
{\tt\small \{felipe.lopes, vasit.sagan\textsuperscript{*}, flavio.esposito\}@slu.edu}
}
\maketitle

\begin{abstract}
Monitoring plantations is crucial for crop management and producing healthy harvests. Unmanned Aerial Vehicles (UAVs) have been used to collect multispectral images that aid in this monitoring. However, given the number of hectares to be monitored and the limitations of flight, plant disease signals become visually clear only in the later stages of plant growth and only if the disease has spread throughout a significant portion of the plantation. This limited amount of relevant data hampers the prediction models, as the algorithms struggle to generalize patterns with unbalanced or unrealistic augmented datasets effectively. To address this issue, we propose PlantPlotGAN, a physics-informed generative model capable of creating synthetic multispectral plot images with realistic vegetation indices. These indices served as a proxy for disease detection and were used to evaluate if our model could help increase the accuracy of prediction models. The results demonstrate that the synthetic imagery generated from PlantPlotGAN outperforms state-of-the-art methods regarding the Fréchet inception distance. Moreover, prediction models achieve higher accuracy metrics when trained with synthetic and original imagery for earlier plant disease detection compared to the training processes based solely on real imagery.
\end{abstract}

\section{Introduction}
\label{sec:intro}
The early detection of plant diseases is crucial for effective monitoring and management during the critical crop growth period. Fortunately, unmanned aerial vehicles (UAVs) have emerged as a valuable tool for efficiently collecting multispectral data and mapping entire fields. Leveraging various spectral bands, these aerial platforms equipped with multispectral cameras provide an opportunity to capture detailed information about plant health. However, training Machine Learning (ML) algorithms for plant health prediction poses a significant challenge due to the limited availability of samples from unhealthy plants. Such data scarcity hampers the development of accurate predictive models as algorithms struggle to generalize patterns effectively with unbalanced datasets. Besides, given the number of hectares to be monitored and flight limitations, plant disease signals become visually clear only at the last stages of plant growth and if the disease has been spread throughout a significant portion of the plantation -- an undesirable end for farmers.

Researchers have attempted to address this challenge, for instance, by detecting the \textit{Puccinia striiformis f.sp. tritici}, commonly known as Wheat Yellow Rust – a devastating fungal disease that affects wheat crops worldwide, leading to significant yield losses if left undetected and untreated. Previous research efforts used traditional data augmentation, such as flipping, cropping, and color jitter, as the most common choices to balance datasets and achieve higher accuracy \cite{zhang2019deep, xu2023detection}. However, these traditional techniques can introduce potential issues, such as a lack of semantic understanding, limited variability, and inadequate preservation of spatial relationships. Recently, new approaches have applied Generative Adversarial Networks (GANs) models to solve similar data augmentation issues. For instance, a previous work \cite{goodfellow2020generative} analyzed object identification with and without different augmentation methods (e.g., flipping, rotation, and translation). The authors concluded that data augmentation and fine-tuned models could improve the identification accuracy and avoid overfitting deep learning networks for tea leaf disease identification with insufficient training set size, but traditional augmentation methods have been unsuccessful in generalization. Our perspective, aligned with the literature \cite{NEURIPS2022_beac6bfb}, is that general GAN models suffer from the lack of semantic understanding and need fine-tunning for generating complex and structured data, such as multispectral imagery from plant plots.

This paper proposes a physics-informed generative adversarial network named PlantPlotGAN to generate more realistic synthetic multispectral imagery focused on plant health analysis from UAV imagery. Its architecture extends DCGAN \cite{radford2015unsupervised}, incorporating physics constraints at the loss function (e.g., reflectance and wavelengths) and balancing the weights according to the most important multispectral bands related to plant disease detection. This provides plant scientists with additional synthetic phenotyping information (including high spatial and temporal resolution).


The main contribution of this work is an innovative physical constraint-based GAN, which contains one generator network and two discriminator networks with different weights that output realistic imagery of plant plots (cf. Figure \ref{fig:synthetic-images}). Such architecture shows that it is possible to extend a GAN to generate higher fidelity imagery for crop analysis, considering the underlying characteristic of reflectance and spectral bands wavelengths. The proposed trained model can provide additional samples to balance a dataset, considering each phase of plant growth. This approach enhances the prediction accuracy for plant disease detection during the early stages of crop development. Other key contributions are:
\begin{itemize}
    \item A latent space manipulation technique to enable the generator to create more realistic multispectral images.
    \item A new layered approach consisting of two discriminators with different responsibilities.
    \item The development of a domain-specific loss function that considers the physics correlation between multispectral bands and the spectral distance between real and fake images.
    \item A robust evaluation model encompassing different perspectives to rank the quality of GAN-based synthetic images.
\end{itemize}

After implementing PlantPlotGAN, we validated our approach by generating synthetic imagery of healthy plots mixed with wheat yellow rust plots and evaluated these generated samples using several metrics, including the Fréchet inception distance (FID) and spectral information divergence (SID). Then, we compared the results with flipped real samples and samples generated from other state-of-the-art GAN architectures.

\begin{figure*}[t]
    \centering
    \includegraphics[width=0.81\linewidth]{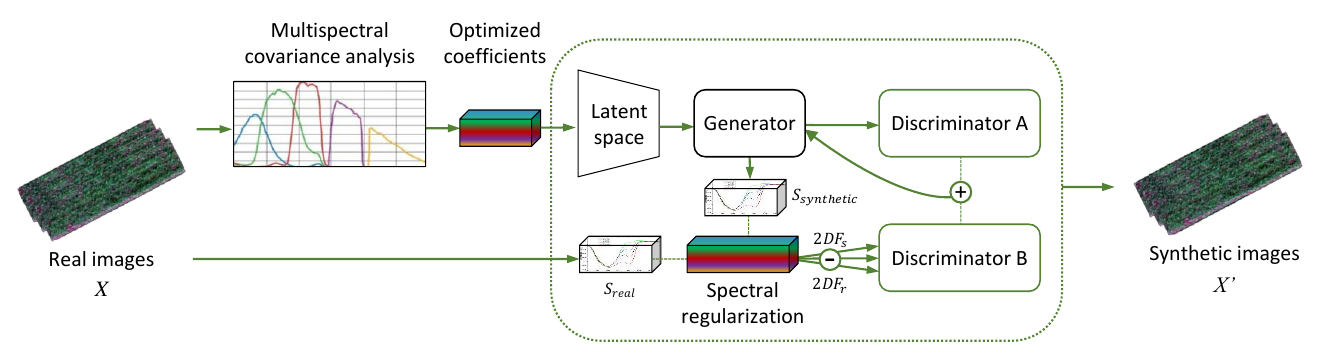}
    \caption{An illustrated overview of the proposed PlantPlotGAN model. According to the optimized coefficients, the physics constraints informing the discriminators and the manipulated latent space are key components to enable an enhanced data augmentation technique.}
    \label{fig:ppgan-overview}
\end{figure*}

\section{Related Work}
Our review of the related work mainly focuses on data augmentation for unbalanced datasets and the generation of synthetic multispectral imagery, the two critical techniques to build and validate the proposed architecture.

\subsection{Traditional Data Augmentation for Limited Imagery Datasets}
Data augmentation is the process of producing more samples from existing data by introducing manipulation techniques, and several research efforts have explored ways to augment data for several distinct areas. For a comprehensive overview of data augmentation techniques, including various GAN-based approaches, interested readers can refer to Khalifa \etal \cite{khalifa2022comprehensive}.

For instance, data augmentation is extensively employed in skin lesion classification, a task that lacks available training data \cite{perez2018data}, while the famous work of Krizhevsky \etal \cite{krizhevsky2017imagenet} applied traditional flipping, rotation, and cropping techniques to augment a dataset for general image classification. In \cite{siddique2022self}, Siddique \etal devised a self-supervised method to segment multiple fruit flower species, utilizing data augmentation to improve the segmentation accuracy. Their method uses a sliding window technique followed by random rotations to increase the variability of samples. Another approach investigated the combination of distinct methods \cite{kaushik2020tomato} (i.e., crop, rotation, and resize) to augment their data, achieving an accuracy of 93\% in detecting tomato leaf disease.

In line with previous research, we used data augmentation. But because of the unique nature of reflectance values used for vegetation indices, we adopted a method that accurately captured sunlight reflection in plants -- to preserve a consistent distribution of multispectral pixel values, different from techniques like flipping or sliding windows that would distort this distribution.


\subsection{GANs and Synthetic Multispectral Imagery}
Although extensive research has explored the use of GAN-based methods to augment data in several domains, e.g., image super-resolution \cite{wang2023multi, liang2022details, demiray2021d}, facial attribute editing \cite{zhang2018generative, huang2023collaborative}, and medical images \cite{shah2022dc, waheed2020covidgan, xue2021selective}, the development of GANs for generating synthetic multispectral imagery is still in its early stages. Specifically within agriculture and remote sensing areas, using multispectral imagery to predict plant disease is paramount due to the valuable information that can be extracted from its bands, such as the vegetation indices \cite{candiago2015evaluating}. Despite the decreasing cost of multispectral sensors, some scenarios (e.g., diseased crops) are rare, limiting the amount of available data.

Fortunately, one of the first papers in generating multispectral satellite images using a generic GAN\cite{abady2020gan} demonstrated success. However, researchers started to give more attention to the lack of a pre-trained model for generating multispectral imagery \cite{bruzzone2019multisource} and the singular covariance between the multispectral bands that hinders the convergence of the GAN training yielding to suboptimal generation \cite{martinez2023ld}.

To the best of our knowledge, \cite{fawakherji2021multi, singh2022data, martinez2023ld} are the closest approaches to the present work in terms of offering GAN-based alternatives to generate synthetic multispectral imagery for predicting plant disease. Martinez \etal \cite{martinez2023ld} employed a pre-trained encoded network and a regularization technique to solve the convergence problem of GANs when trained with multispectral images due to their high dimensionality. Fawakherji \etal \cite{fawakherji2021multi} used a conditional GAN (cGAN) to generate RGB data with near-infrared (NIR) information, generating four-channel multispectral synthetic images for weed recognition tasks. In \cite{singh2022data}, the authors incorporated the spectral behavior of land-use and land-cover classes into a GAN to better model the properties of classes by using spectral indices.

Although the similar fact of using GAN to generate multispectral imagery, this proposal distinguishes itself from previous endeavors by adding a new discriminator to verify physics constraints into such generation, and assessing the spectral similarity between real and synthetic imagery. This paper also analyzes the advantages of employing synthetic data to improve machine learning algorithms' accuracy, but focusing on the early detection of plant diseases.

\section{Methodology}

\subsection{Problem Formulation}
In this section, we formalize the problem of generating multispectral synthetic imagery. This encompasses the definition of the optimizer used to acquire physically constrained spectral profiles.

As depicted in Figure \ref{fig:ppgan-overview}, $X$ and $X'$ respectively denote the set of real input multispectral images and the set of possible synthetic multispectral images. In multispectral images, especially in the context of remote sensing and vegetation analysis, there is a covariance between the last two spectral bands, i.e., Near Infrared (NIR) and Red-Edge (RE). Then, if there is a covariance between the last two bands of $X$, $X'$ should reproduce the same covariate behavior in its NIR and RE pixel values (which are the inputs of our covariance calculation). This technique forces the randomness of latent space to follow the existing physical constraint in the relation between NIR and RE \cite{penuelas1998visible}.
For each input/source image $x$ belonging to $X$, its covariance coefficients $c$ should also be in any $x'$ element belonging to $X'$. Given a set of coefficients $C$ and an input $X$, the goal of a generative multispectral model is to train a generator denoted as $G(z; \theta_h)$, where $z$ is the input noise vector and $\theta_h$ denotes the generator's parameters. The goal of $G$ is to generate the elements of $X'$, using the coefficients of $C$ and observing the evaluation of discriminators $D$. For each generated $X'$, two discriminators denoted by $D_1(X; X'; \theta_1)$ and $D_2(X_{\text{spectral}}; X'_{\text{spectral}}; \theta_2)$ are responsible for evaluating the distribution and covariance of $X'$ in relation to $X$, respectively, where $X_{\text{spectral}}$ and $X'_{\text{spectral}}$ refer to the spectral profile of $X$ and $X'$, and $\theta_1$ and $\theta_2$ denote their parameters.

\subsection{PlantPlotGAN Model}
\label{sec:ppgan-model}
The model proposed in this paper consists of five main modules: optimizer, spectral regularization, generator, and two discriminators, as depicted in Figure \ref{fig:ppgan-overview}. To facilitate the comprehension of our PlantPlotGAN model as well for notation simplicity, consider $X$ and $X'$ vectors $(N, N, 5)$, in the sense that 5-band multispectral imagery will be considered, and $N$ is the imagery size -- note that although the plant plots are rectangular, we assume square imagery for the sake of simplicity in the convolution layers of each GAN model and consider only the inner rectangle in each analysis (e.g., spectral profile, FID). The architecture of PlantPlotGAN is an extension of DCGAN \cite{radford2015unsupervised}, adding a new discriminator to validate the spectral profiles of synthetic imagery and can be incremented to handle other physical constraints.
\\
\\
\noindent\textbf{Optimizer.} Using the elements of $X$, the optimizer denoted by $O$ is responsible for obtaining spectral coefficients, which manipulate the latent space. To achieve that, the optimizer selects the NIR and RE bands of $X$, obtains the covariance $Cov$, and minimizes a function to return three coefficients that approximate the equation \ref{eq:rededge} to $Cov$. The selection of NIR and RE bands is based on the scientific observation that vegetation exhibits distinctive spectral properties, particularly in the RE and NIR regions of the electromagnetic spectrum \cite{filella1994red}. 

In Equation \ref{eq:rededge}, $RE(\lambda)$ represents the Red Edge channel reflectance at a given wavelength $\lambda$, and $NIR(\lambda)$ represents the Near Infrared channel reflectance at the same wavelength. The parameters $G$, $H$, and $K$ are the coefficients that control the shape, characteristics, and covariance between the Red Edge and Near Infrared channels, respectively. The latent space, then, is weighted according to the coefficients adjusted during the convergence process.

\begin{equation}
RE(\lambda) = G \cdot e^{-H \cdot \lambda} + K \cdot NIR(\lambda)
\label{eq:rededge}
\end{equation}

\noindent\textbf{Spectral Regularization}. Formally, given an input $X$, the generator $G$ should generate a set $X'$ with a similar spectral profile. However, depending on the number of elements $x \in X$, $G$ can generate elements $x'$ with an abnormal spectral profile. It is thus necessary to conduct factorization on $X'$. The Spectral Regularization module, denoted as $SR(S_{x},S_{x'})$, calculates the spectrum of $x$ and $x'$, using an underlying 2D Fast Fourier Transform (FFT) to compute and compare the shift-invariance and noise of each set. The result of $SR$ is in one of the discriminators as a spectral loss.
\\
\\
\noindent\textbf{Generator.} Our generator $G$ receives a random noise $n$, with the latent space values already adjusted by the optimizer $O$ as input, and generates the output synthetic imagery $x'$. The objective of $G$ is to generate samples that fool the discriminators into classifying $x'$ as real. As in DCGAN \cite{radford2015unsupervised}, $G$ utilizes convolutional layers to transform $n$ into a synthetic $x'$. After receiving the random noise from the optimized latent space, $G$ upsamples vector $n$ into a higher-dimensional representation using deconvolutional layers. PlantPlotGAN has three deconvolutional layers and LeakyReLU activation functions in sequence to introduce non-linearity in data generation.
\\
\\
\noindent\textbf{Discriminators.} For each set of synthetic samples $X'$, two discriminators are responsible for i) evaluating how close to $x$ each element $x' \in X'$ is; and ii) calculating the spectral divergence between each $x$ and $x'$. For the first objective, our PlantPlogGAN utilizes the discriminator $D_1(x; x'; \theta_s)$ to compute the probability that $x'$ comes from the real data distribution rather than $G$. The second discriminator, $D_2(x; x'; \theta)$, receives the spectrum of $x$ and $x'$ and calculates a dynamic distance to identify if the spectrum $x_i$ belongs to a spectral profile of $x$.
\\
\\
Thus, the objective of PlantPlotGAN with one generator and two discriminators is to find equilibrium by solving the following minimax problem:

\begin{align}
\min_h \max_s &V(D_1, D_2, G) = \notag \\
& E_{x \sim p_{\text{data}}(x)} [\log D_1(x) + \log D_2(x)] + \notag \\
& E_{z \sim p_z(z)} [\log (1 - D_1(G(z))) + \notag \\
& \hspace{4em} \log (1 - D_2(G(z)))],
\end{align}

\noindent where ${E}_{x \sim p_{\text{data}}(x)}$ denotes the expectation over real data samples. ${E}_{z \sim p_z(z)}$ refers to the expectation over the gradients for the spectral vector sampled from a prior distribution $p_z(z)$. $D_1(x)$ represents the output probability of the first discriminator for real data $x$. $D_2(x)$ represents the output probability of the second discriminator for real data $x$. $D_1(G(z))$ is the output probability of the first discriminator for synthetic data generated by the generator, while $D_2(G(z))$ represents the output probability of the second discriminator for synthetic data generated by the generator. Finally, $V(D_1, D_2, G)$ is the value function of the \textit{minimax} game. It is maximized with respect to the discriminator parameters ($\theta_1 \text{ and } \theta_2$) and minimized with respect to the generator parameters $\theta_h$.
Similar to other state-of-the-art GANs \cite{goodfellow2020generative}, PlantPlotGAN updates the parameters of $G$ and $D$ iteratively to improve their performance.

\subsection{Dataset Preparation}
\label{sec:dataset-prep}
To train the convolution networks $G$ and $D$, and validate the PlantPlotGAN architecture for its aim, a new multispectral imagery dataset of spring wheat was collected at Chacabuco (-34.64; -60.46), Argentina. The data collection experiment consisted of growing three varieties of spring wheat, categorized as an intermediate-ate-long cycle, intermediate cycle, and short cycle, in a total of 700 field plots. The plots were arranged in a grid pattern with 35 rows and 20 columns within the field. Each plot had dimensions of 1.2 x 5 meters and consisted of 7 rows of wheat plants. Spring wheat was planted on June 19th, 2021, and harvested on December 16th, 2021. The soil type at the site was classified as fine-silty, mixed, thermic Typic Argiudoll according to the USDA-Soil Taxonomy V. 2006.

Throughout the growing season, uniform field management practices were implemented for all field plots. Insecticide, herbicide, and fertilizer were applied as needed to ensure consistent plant health and growth. However, no fungicide was used in the experiment, as the focus was on studying the impact of the biotic stress caused by the occurrence of yellow rust disease.
\\
\begin{figure*}[htbp] 
  \centering
  \includegraphics[width=0.51\linewidth]{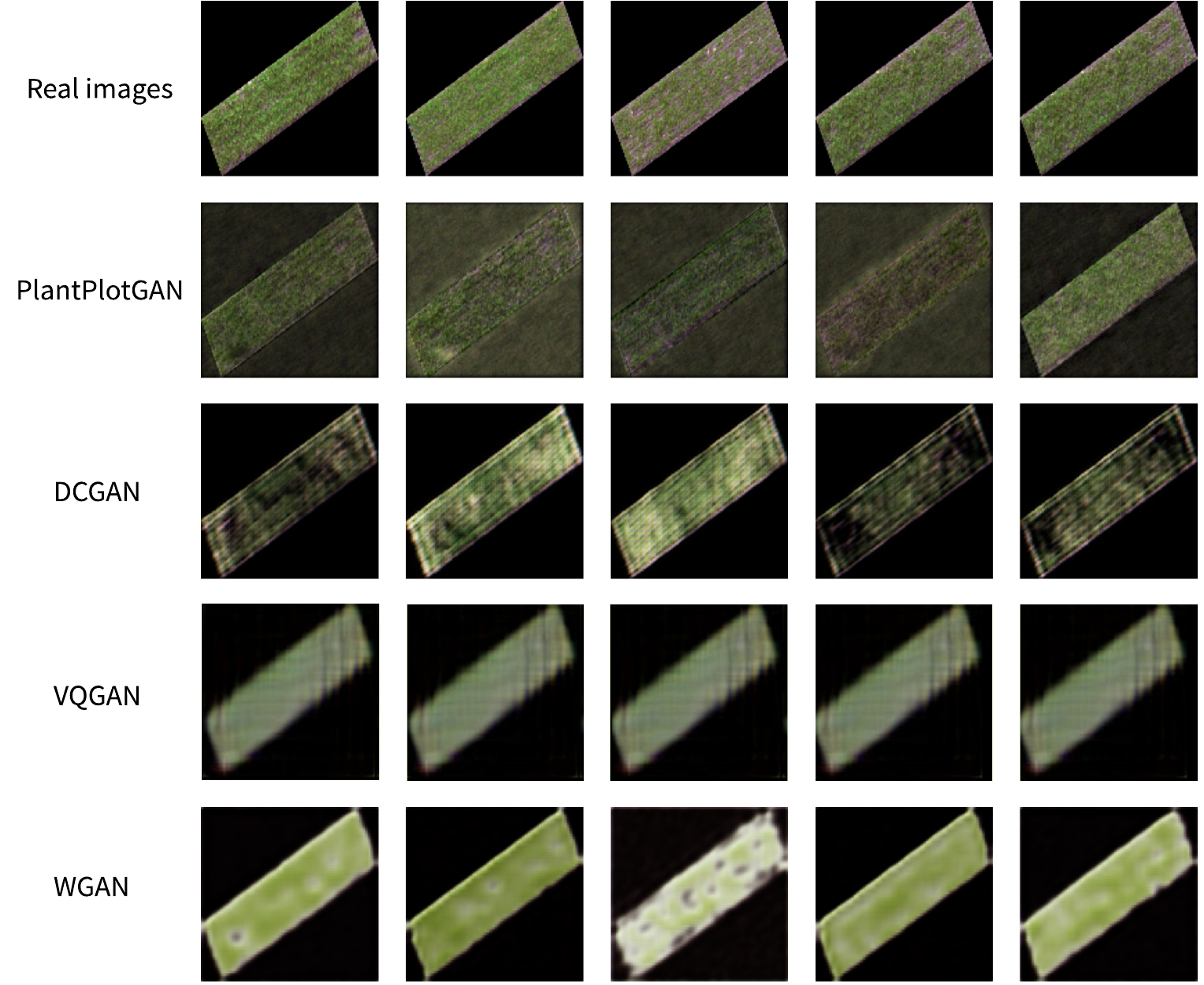}
  \caption{Results of each evaluated GAN model after 50 epochs and 5 training steps. Despite the valuable information for accurate plant health analysis, the high similarity between pixels in vegetation multispectral imagery makes converging unspecialized models harder.}
  \label{fig:synthetic-images}
\end{figure*}

\noindent\textbf{UAV flights.} The primary aerial platform for collecting remotely sensed data from the standing crop was the DJI Phantom 4 Pro. This UAV was equipped with a multispectral sensor capable of capturing the five discrete spectral bands used with the PlantPlotGAN modules (cf. Section \ref{sec:ppgan-model}): blue, red, green, red edge, and near-infrared. A sunlight irradiance watcher was connected to the sensor to ensure accurate measurements. To achieve a high spatial resolution of the imagery data, the UAV was programmed to fly at a low altitude of 20 meters from the surface, resulting in a pixel resolution of 1.04 centimeters. To monitor the crop at different stages of growth, the aerial data collection was designed to have a temporal dimension. The UAV was flown five times throughout the growing stages, starting from the first flight on August 30th, 2021, during the tillering stage, and concluding with the last flight on November 17th, 2021, during the flowering stage.
\\
\\
\noindent\textbf{Ground truth data.} On November 15th, 2021, a team of experts, including pathological and phenological specialists, visited the experiment site to assess the wheat plants' health status visually. Their primary task was to provide annotations and divide the plants into different categories based on their health condition: healthy, average, and severe disease. The specialists determined the level of disease infection by carefully examining the leaves of the plants in each plot. The entire plot was labeled as healthy if the leaves showed no signs of green loss. The plot was labeled as having a mild infection if some yellow spots were present. Conversely, if most plants in the plot exhibited a yellowish color, it was labeled as having a severe infection.

It is important to note that due to labor constraints, the crop health status was assessed and annotated for 592 out of the 700 plots. Of these 592 samples, 430 (72.6\%) were classified as mild, 106 (17.9\%) as unhealthy, and 56 (9.45\%) as healthy. These quantities of samples for each class demonstrate the common imbalance scenario of real-world data and its related challenges for an accurate prediction \cite{kaur2019systematic}.



\section{Experiments}
The validation of the proposed method used two sets of experiments and two related datasets. The first group evaluates image quality metrics regarding the synthetic images generated by PlantPlotGAN and compares them to the output of other state-of-the-art models, i.e., DCGAN \cite{radford2015unsupervised}, WGAN \cite{arjovsky2017wasserstein}, and VQGAN \cite{esser2021taming}. Besides, a quantitative performance evaluation is part of this first group. The second group of experiments utilizes the synthetic images generated from PlantPlogGAN to extract vegetation indices and assess whether it could improve the accuracy metrics of machine-learning models in detecting wheat yellow rust.

\subsection{Dataset and Data Extraction}
\label{sec:dataset}
The dataset used in both sets of experiments considers the data presented in Sub-section \ref{sec:dataset-prep}. For the training of PlantPlotGAN, its modules (i.e., discriminators, optimizer, and spectral regularizer) used only the labeled samples as healthy or unhealthy while ignoring the mild samples, as their mixed signals would hinder convergence. For fairness in the evaluation and to reduce training costs, the remaining 162 images were resized to 128x128x5. The quality assessment and comparison of PlantPlotGAN with other models used the same dataset in all scenarios and metrics.

In the set of experiments for yellow rust prediction, a vegetation index extractor using the Rasterio library derived 47 indices for each sample, such as the Normalized Difference Vegetation Index (NDVI), Improved Modified Chlorophyll Absorption Ratio Index (MCARI), and Green Chlorophyll Index (GCI). The same process was used to extract these vegetation indices from the synthetically generated samples. The choice for these 47 vegetation indices aligns with the literature \cite{croft2014applicability}.

\subsection{Implementation}
During training, all models use the Adam \cite{kingma2014adam} optimizer with the learning rate set to 2e-4, $\beta1 = 0.5$, and $\beta2 = 0.99$. The observations are collected every 5 steps in 50 epochs, finalizing with the trained generators. Figure \ref{fig:synthetic-images} depicts some of the synthetic samples created by the generators. Notably, even the VQGAN model, equipped with multiple underlying convolutional layers (pre-trained using VGG16) and encoders, fell short of generating satisfactory results that visually resembled real images within the defined number of epochs and training images.

The generator and discriminator networks utilize 2D transpose and 2D convolution layers for each GAN model. Each subnetwork incorporates batch normalization and LeakyReLU activations following the convolutional layer, as described in \cite{radford2015unsupervised}. For all GAN experiments, a random input $z \in \mathbb{R}^{100} \sim \mathcal{N}(0, 1)$ was used as initial noise. Every implemented GAN model uses the TensorFlow platform and runs in a single Intel Core i9-10900X CPU @ 3.70Ghz and 128 GB. Training the PlantPlotGAN and the other three GAN models on all scenarios takes about 14 hours. More details are given in the supplementary material.



\subsection{Performance Comparision}
Using the same number of epochs (50) and training steps (5) for all the evaluated GANs, the convergence process of each model was assessed by continuously observing the adversarial loss of the discriminators and generators. Subsequently, the metrics described in the following sub-sections were collected and compared after the training phase.

\subsubsection{Similarity, Fidelity, and Spectral Analysis}
\label{sec:metrics}
An in-depth analysis compared the real samples with the synthetic dataset generated by DCGAN, WGAN, VQGAN, and PlantPlotGAN to assess the quality of synthetic samples. The objective was to assess the similarity and fidelity of the synthetic imagery. The following metrics were employed in this analysis:
\\
\\
\noindent\textbf{Fréchet Inception Distance (FID).}
The FID is a widely used metric for evaluating the quality of imagery generated from GAN models. It measures the similarity between real and generated images' distribution by utilizing features extracted from a pre-trained Inception-v3 neural network. The FID quantifies the dissimilarity between the two distributions in a feature space. A lower FID score indicates a higher similarity between the real and generated images, implying better quality and fidelity in the generated imagery. By considering both the distributional and perceptual aspects of the images, the FID offers a comprehensive evaluation of the GAN model's image generation capabilities. FID is denoted by:

\begin{align}
\text{FID} = &\|\mu_{\text{real}} - \mu_{\text{generated}}\|^2 
+ \text{Tr}(\Sigma_{\text{real}} + \Sigma_{\text{generated}} \notag \\ 
&- 2(\Sigma_{\text{real}}\Sigma_{\text{generated}})^{1/2})
\end{align}

\noindent where $\mu_{\text{real}}$ and $\mu_{\text{generated}}$ are the mean vectors of features extracted from the real and generated images, respectively. $\Sigma_{\text{real}}$ and $\Sigma_{\text{generated}}$ represent the covariance matrices of the feature distributions. The FID captures both the difference in means and the difference in covariances between the real and generated image feature distributions, providing a quantitative measure of their dissimilarity. For our scenario, as Inception-v3 cannot handle multispectral images, a mean of two FID calculations is considered. The first calculation selected bands 1:3, while the second computation selected bands 3:5 from both datasets (i.e., real and synthetic).
\\
\\
\noindent\textbf{Chi-square.} It was used to compare the observed and expected frequencies within each spectral band, enabling the detection of significant deviations. The equation to obtain this metric is denoted by:

\begin{align}
\chi^2 = \sum \frac{{(O_i - E_i)^2}}{{E_i}},
\end{align}

\noindent where $\chi^2$ represents the Chi-square test statistic, $O_i$ denotes the observed frequency of a particular category or bin in the dataset, and $E_i$ represents the expected frequency based on a specified hypothesis or model. We can determine whether the observed frequencies significantly differ from the expected frequencies by comparing the computed Chi-square value to the critical Chi-square value corresponding to a chosen significance level and degrees of freedom. A higher Chi-square value indicates a larger deviation between the real images and the synthetic data, suggesting a higher dissimilarity or lack of agreement. 
\\
\\
\noindent\textbf{Intersection and Bhattacharyya.} The Chi-square method alone may not capture the complete picture of similarity or dissimilarity between the two sets of images. Intersection and Bhattacharyya are complementary coefficients to achieve a deeper analysis. The Intersection coefficient (IC) considers the overlap or commonality between the pixel values of real and synthetic images. It calculates the ratio of the intersection of pixel values to the union of pixel values in the two datasets. By examining the shared elements between the sets, the IC reveals the extent of similarity or overlap between the real and synthetic images. The Bhattacharyya coefficient (BC) quantifies the similarity between the probability distributions of the real and synthetic datasets. It can be computed as the Bhattacharyya distance, which measures the divergence between the two distributions. IC and BC are denoted respectively by the following equations:

\begin{align}
\label{eq:intersection}
\text{IC} = \sum_{i=1}^{n} \min(P_{\text{real}_i}, P_{\text{synthetic}_i}), \text{and}
\end{align}
\begin{align}
\label{eq:bc}
\text{BC} = \sum_{i=1}^{n} \sqrt{P_{\text{real}_i} \cdot P_{\text{synthetic}_i}},
\end{align}

\noindent where \(P_{\text{real}_i}\) and \(P_{\text{synthetic}_i}\) represent the probability densities of the real and synthetic datasets at the \(i\)-th spectral band, respectively. The sums are taken over all spectral bands (from \(i = 1\) to \(n = 5\)) for the experiments.

A higher Intersection coefficient indicates a high match or complete overlap, implying a higher similarity between the real and synthetic images. Conversely, a lower value suggests a lower similarity between the sets. On the other hand, generating imagery with a lower BC is desirable for every GAN model.

\subsubsection{Evaluation of Early Plant Disease Detection}
Considering the potential of our collected data, by having a temporal dimension, we also investigated if using the synthetic data generated from PlantPlotGAN could improve early detection of the \textit{Puccinia striiformis f.sp. trit-ici}, i.e.,  wheat yellow rust. A feature-based prediction was employed using classical machine learning algorithms. The algorithms were trained with two datasets, i.e., real images and real images mixed with synthetic samples. The test dataset consisted of real images only. The metrics used to evaluate the early prediction are accuracy, recall, and F1 Score. We selected these metrics due to the cost of false positives (incorrectly labeled negatives) for farmers, i.e., a crop classified by mistake as diseased would receive toxic treatment unnecessarily. The other side is equally costly; a false negative classification could lead to truly diseased crops being classified as healthy and left without treatment.

\section{Results\protect\footnote{Additional evaluations, results, and the trained models can be found in the supplementary material.}}

\begin{figure}[t]
    \centering
    \includegraphics[width=0.93\linewidth]{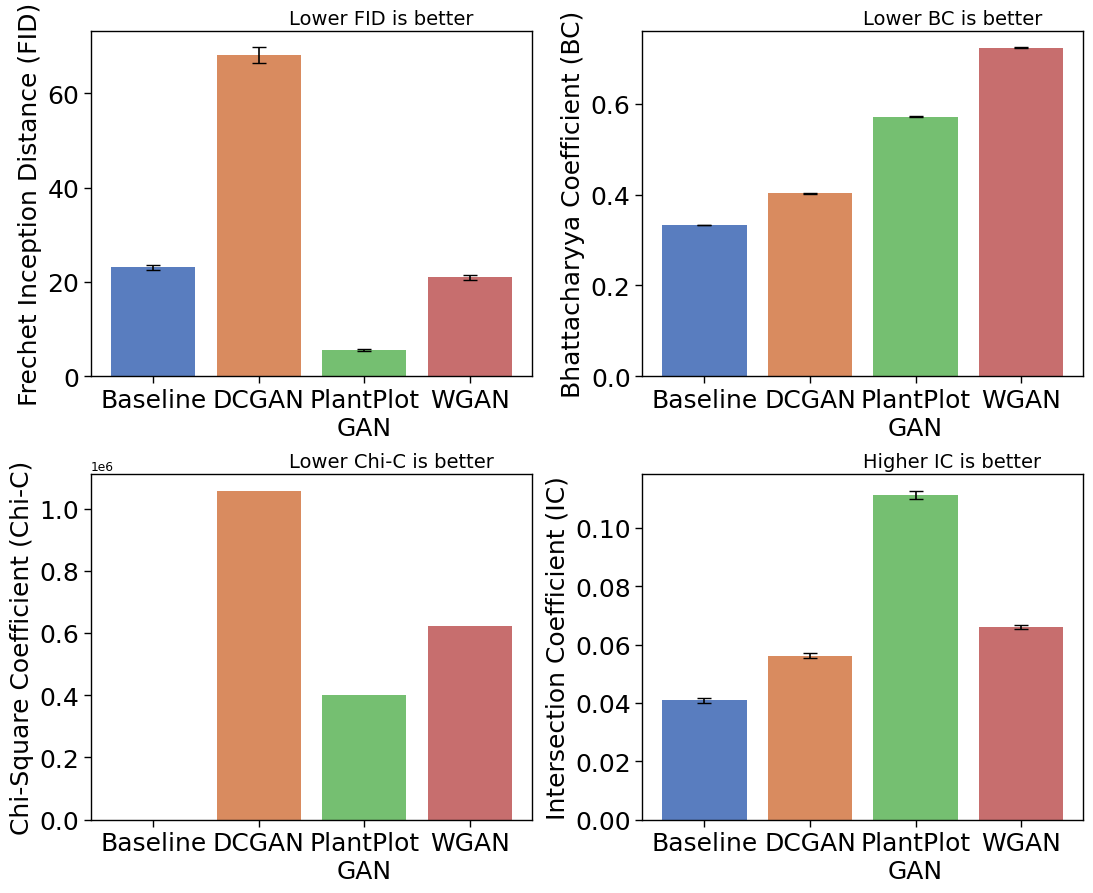}
    \caption{Analyzing each metric discussed in Section \ref{sec:metrics}. PlantPlotGAN achieves improved results in almost all metrics.}
    \label{fig:chart-metrics}
\end{figure}

\noindent\textbf{Similarity, Fidelity, and Spectral Analysis.} As shown in Section \ref{sec:metrics}, the first set of experiments evaluated the quality of the generated images. To this aim, the evaluation considered four metrics (i.e., FID, Chi-square, IC, and BC). Figure \ref{fig:chart-metrics} depicts the result of this evaluation. The baseline refers to the flipped real images.

Particularly, our PlantPlotGAN leads to significant improvements compared to the other GAN architectures in three evaluation metrics. PlantPlotGAN achieved the best average FID score, indicating that it can generate images with the highest quality. It also achieved a higher IC score, meaning a more precise distribution of pixels along the generated imagery. 
These metrics help clarify the result in Figure \ref{fig:synthetic-images}, demonstrating that an analysis solely based on FID may not represent the true quality of synthetic data generated from GAN models. For instance, DCGAN generated synthetic images visually closer to the real vegetation plots for the same number of epochs. However, WGAN obtained a lower FID than DCGAN for this scenario.

\begin{figure}[t]
    \centering
    \includegraphics[width=0.93\linewidth]{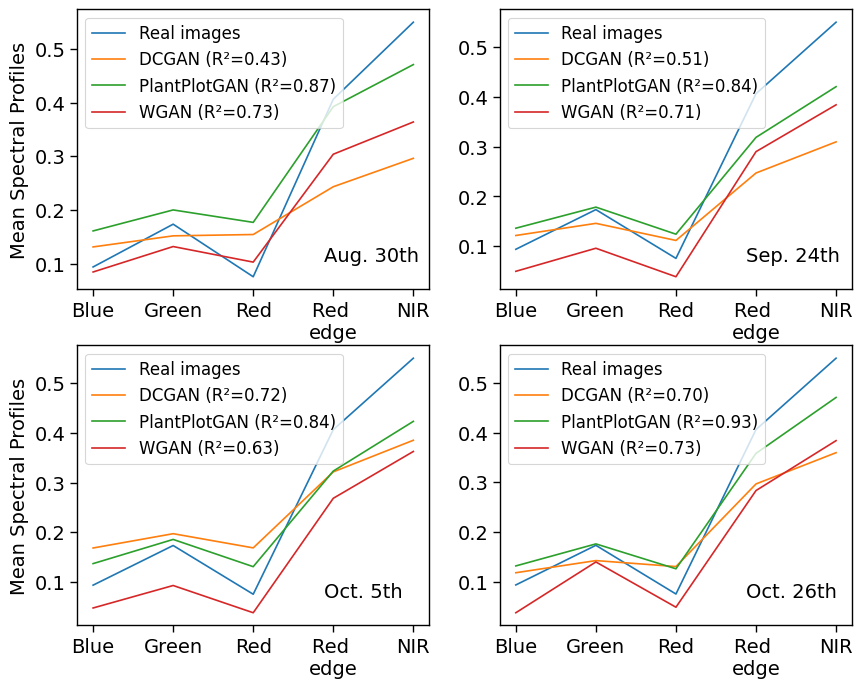}
    \caption{Comparison of mean spectral profiles between real and synthetic imagery, encompassing all evaluated GAN models and the growing stage dates from the dataset discussed in Section \ref{sec:dataset}.}
    \label{fig:chart-mean-spectral-profile}
\end{figure}

\begin{table}[]
\centering
\resizebox{\columnwidth}{!}{%
\begin{tabular}{c|cccccc}
\hline
\hline
\multirow{2}{*}{Metrics} & \multicolumn{2}{c|}{XGboost}                        & \multicolumn{2}{c|}{Random Forest}          & \multicolumn{2}{c}{CNN} \\
                         & Real          & \multicolumn{1}{c|}{Real+Synthetic} & Real  & \multicolumn{1}{c|}{Real+Synthetic} & Real   & Real+Synthetic \\ \hline\hline
F1 Score                  & 0.69 & \multicolumn{1}{c|}{0.821}         & 0.678 & \multicolumn{1}{c|}{0.732} & 0.8   & 0.841 \\ \cline{1-1}
Accuracy                 & 0.72          & \multicolumn{1}{c|}{0.821}          & 0.678 & \multicolumn{1}{c|}{0.732}          & 0.808  & \textbf{0.842} \\ \hline
\multirow{4}{*}{Accuracy} & \multicolumn{6}{c}{\textit{healthy}}                                                           \\ \cline{2-7} 
                          & 0.73 & \multicolumn{1}{c|}{0.864}         & 0.8   & \multicolumn{1}{c|}{0.8}   & 0.814 & 0.83  \\ \cline{2-7} 
                          & \multicolumn{6}{c}{\textit{unhealthy}}                                                         \\ \cline{2-7} 
                         & \textit{0.69} & \multicolumn{1}{c|}{0.74}           & 0.52  & \multicolumn{1}{c|}{0.6}            & 0.789  & \textbf{0.825} \\ \hline
\multirow{4}{*}{Recall}   & \multicolumn{6}{c}{\textit{healthy}}                                                           \\ \cline{2-7} 
                          & 0.91 & \multicolumn{1}{c|}{0.864}         & 0.67  & \multicolumn{1}{c|}{0.781} & 0.916 & 0.92  \\ \cline{2-7} 
                          & \multicolumn{6}{c}{\textit{unhealthy}}                                                         \\ \cline{2-7} 
                          & 0.36 & \multicolumn{1}{c|}{\textbf{0.74}} & 0.68  & \multicolumn{1}{c|}{0.64}  & 0.6   & 0.66  \\ \hline
\multirow{4}{*}{F1 Score} & \multicolumn{6}{c}{\textit{healthy}}                                                           \\ \cline{2-7} 
                          & 0.81 & \multicolumn{1}{c|}{0.86}          & 0.73  & \multicolumn{1}{c|}{0.79}  & 0.862 & 0.88  \\ \cline{2-7} 
                          & \multicolumn{6}{c}{\textit{unhealthy}}                                                         \\ \cline{2-7} 
                          & 0.47 & \multicolumn{1}{c|}{\textbf{0.74}} & 0.59  & \multicolumn{1}{c|}{0.62}  & 0.681 & 0.733 \\ \hline\hline
\end{tabular}%
}
\caption{Comparing the \textit{Real} and \textit{Real+Synthetic} data, we can observe that the \textit{Real+Synthetic} values are generally higher across the different models and metrics. This suggests that incorporating synthetic data into the training process has a positive impact.}
\label{table:scores}
\end{table}

\begin{figure}
    \centering
    \includegraphics[width=0.9\linewidth]{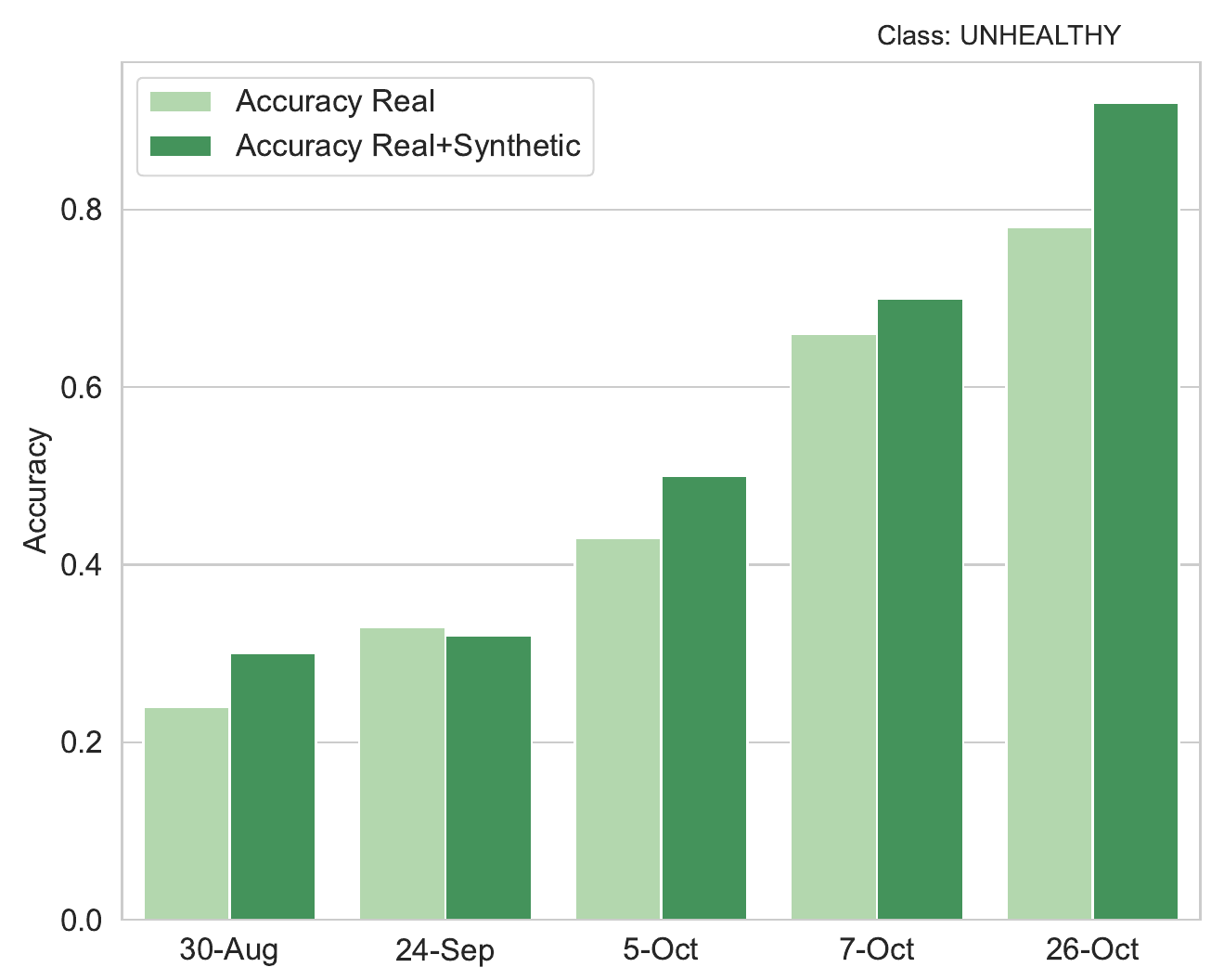}
    \caption{Evaluation of a combined dataset containing real and synthetic images for detecting yellow rust disease. The analysis considers the temporal progression, where each subsequent date considers the data acquired from previous dates.}
    \label{fig:chart-time-series}
\end{figure}

The quality of synthetic images was also measured by the spectral profile shape -- as it is the source of several vegetation indices used to asses plant health (cf. Section \ref{sec:dataset}). VQGAN was excluded from this analysis due to the current limitation on generating only RGB images. In this case, as depicted in Figure \ref{fig:chart-mean-spectral-profile}, our method achieved a satisfying $R^2$ in all scenarios compared to the spectral profile of real images. It also demonstrates the spectrum regularizer $SR$ in action, weighting the loss of $D_2$ to approximate the model function. Future work can assess whether a simple increase in the number of parameters of traditional GAN models could achieve the same result obtained by PlantPlotGAN. Besides, the implementation evaluated in this experiment only considered the covariance between sequential channels (e.g., Red edge and NIR). Extending the $SR$ to perform a combinatorial covariance analysis of multispectral bands is an open issue.

\noindent\textbf{Early plant disease detection}
The second set of experiments evaluated the potential of PlantPlotGAN to improve the early detection of yellow rust. After excluding the mild samples (as discussed in Section \ref{sec:dataset}), the dataset with real imagery has 106 healthy samples (65.43\%) and 56 unhealthy samples (34.56\%). To assess if PlantPlotGAN could overcome the imbalance problem, the first experiment compared a prediction model trained in two scenarios for the extracted vegetation indices i) extraction from the real imagery data; ii) extraction from the mixed dataset consisting of real and synthetic imagery. The validation dataset consists of a stratified selection of the real imagery at the last observed date (cf. Section \ref{sec:dataset}).
PlantPlotGAN generated 50 healthy and 10 unhealthy samples for the mixed dataset, resulting in a distribution of 116 healthy samples (52.2\%) and 106 unhealthy samples (47.74\%). For obtaining the accuracy indices, three models were utilized, i.e., XGBoost, Random Forest, and a Convolutional Neural Network (CNN). The description of the parameters of each model is present in the supplementary material. 
This first evaluation measured if the accuracy scores would increase when utilizing the combined dataset for training. Table \ref{table:scores} demonstrates the observed results, with the combined PlantPlotGAN-based dataset achieving enhanced values in almost all scenarios.
Furthermore, a per-date analysis verified if the F1 Score would increase by detecting the \emph{unhealthy} class at early dates for the evaluated scenario. Figure \ref{fig:chart-time-series} depicts the results of this experiment, demonstrating an overall detection improvement of employing the synthetic data generated from PlantPlotGAN in detecting yellow rust disease at early stages.

\section{Conclusion}
We have proposed a novel generative adversarial network model based on latent space manipulation and an architecture of two discriminators. The proposed PlantPlotGAN is the first end-to-end GAN model for generating multispectral plant plots and is effective for augmenting UAV-based plant imagery datasets. This is due to two key components: a novel spectrum regularizer that factories the latent space with optimized spectral coefficients; and a two-layered discriminator architecture that utilizes physics constraints to decide if a spectrum is true or fake.
Extensive experiments show that our PlantPlotGAN achieves significant improvements over state-of-the-art GAN models and highlights the importance of incorporating physical constraints for synthetic data generation.

In future work, a per-channel discriminator might improve the synthetic imagery quality. Besides, incorporating a style transfer based on StyleGAN \cite{karras2020analyzing} can be a fruitful approach to take advantage of a full categoric multispectral dataset (e.g., diseased plants). 
{\small
\bibliographystyle{ieee_fullname}
\bibliography{egbib}
}

\end{document}